\documentclass[journal]{IEEEtran}
\usepackage{framed}
\usepackage[ruled]{algorithm2e}
\usepackage{amssymb}
\usepackage{latexsym}
\usepackage{epsfig}
\usepackage{float}
\usepackage{graphicx}
\usepackage{booktabs,multirow, multicol}
\usepackage{cite}
\usepackage{epsfig}
\usepackage{caption}
\usepackage{rotating}
\usepackage{epstopdf}

\usepackage{tabularx}
\usepackage{url}
\usepackage{xcolor}
\definecolor{newcolor}{rgb}{.8,.349,.1}

\graphicspath{{pics/}}
\begin{document}

\title{ESCaF: Pupil Centre Localization Algorithm with Candidate Filtering}

\author{Anjith~George,~\IEEEmembership{Member,~IEEE,}
        and~Aurobinda~Routray,~\IEEEmembership{Member,~IEEE}
\thanks{ Thanks!}}

\maketitle

\begin{abstract}
Algorithms for accurate localization of pupil centre is essential for gaze tracking in real world conditions. Most of the algorithms fail in real world conditions like illumination variations, contact lenses, glasses, eye makeup, motion blur, noise, etc. We propose a new algorithm which improves the detection rate in real world conditions. The proposed algorithm uses both edges as well as intensity information along with a candidate filtering approach to identify the best pupil candidate. A simple tracking scheme has also been added which improves the processing speed. The algorithm has been evaluated in Labelled Pupil in the Wild  (LPW) dataset, largest in its class which contains real world conditions. The proposed algorithm outperformed the state of the art algorithms while achieving real-time performance.
\end{abstract}

\section{Introduction}

Head-mounted eye trackers are useful in investigating human behavior in
many practical dynamic tasks. In head-mounted cameras, accurate localization of
pupil center is possible with the use of near-infrared (NIR) illumination, which can give improved accuracy. Algorithms for accurate localization of pupil centre is essential for gaze tracking in real-world conditions. Iris-sclera boundary is prominent in visible image based gaze tracking, where as in NIR lighting the pupil-iris boundary is much more highlighted. Most of the head-mounted eye trackers utilize dark pupil method for localizing the pupil. Even though this method superior in indoor, controlled conditions, the accuracy degrades in challenging outdoor conditions. As head-mounted eye trackers are supposed to function in challenging outdoor conditions, the pupil center detection algorithm should be robust against real-world conditions.

 Most of the existing algorithms for pupil localization perform well only under controlled conditions. With the advent of wearable head-mounted devices \cite{starner2013project},\cite{benko2015fovear},  eye tracking holds the potential to become a human-computer interaction channel. The point of gaze gives a lot of information regarding the attention of the user, which can be used to manipulate objects in virtual and augmented reality environments. Gaze tracking can also be used for foveated rendering \cite{guenter2012foveated}, which can reduce the computational load in image rendering. Head-mounted trackers have more potential as they are not limited to desktop environments. However, the accuracy of gaze tracking degrades with real-world conditions such as illumination variations, blur, partial-occlusions, reflections from external light sources, makeup, contact lenses,  and other sources of noise.  Therefore a robust pupil localization algorithm is necessary to deal with such real-world conditions. 

\begin{figure}[h]
\begin{center}
\includegraphics[width=1\linewidth]{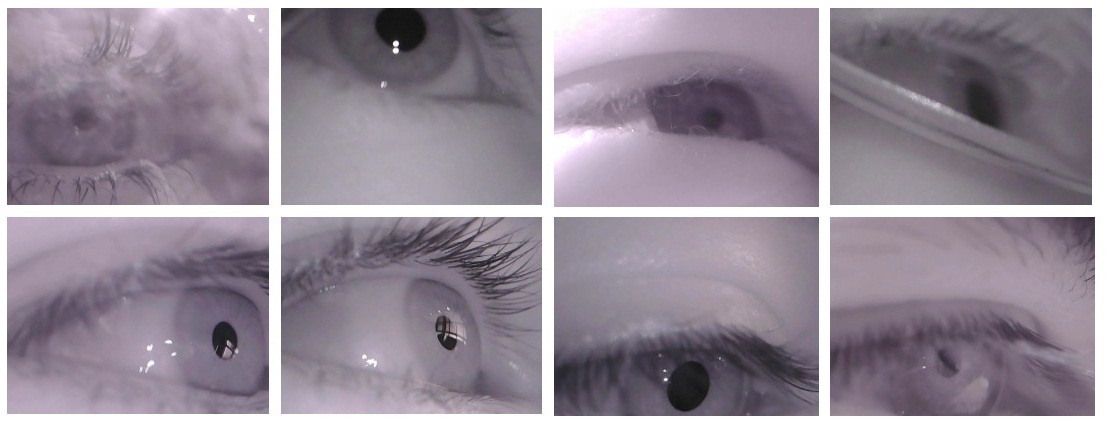}
\end{center}
\caption{Sample images from the LPW dataset}
\label{fig:samples_lpw}
\end{figure}

Pupil localization can be performed easily if the following conditions are true. 1) The gradients of pupil boundary are strong and can be detected by an edge detector 2) Pupil is the darkest region in the image. However, these assumptions may not hold well in uncontrolled environments. In this work, we develop a hybrid approach which can robustly detect PC in the images obtained from a head-mounted camera. The algorithm can be further extended to work with remote eye trackers by adding an eye detection stage. The algorithm uses multiple features and fits an ellipse with a candidate filtering scheme. The proposed approach is named as \textit{Ellipse Selection by Candidate Filtering} (ESCaF). Dark pupil images obtained from a head-mounted camera are shown in Fig. \ref{fig:samples_lpw}.

A hybrid approach is proposed which uses intensity distribution as well as the edges for PC localization. Additionally, a simple tracking scheme is added which increases the detection rate in real-world conditions. The focus of the paper is on developing an algorithm for a head-mount eye tracker without the requirement of specially designed illuminators and optics. The algorithm designed to work even with off the shelf webcams and infrared LEDs. 
 
The main contributions from this paper are listed below.

\begin{itemize}

\item Proposes a novel framework for pupil center localization in NIR (Near Infrared) images.
\item The proposed approach combines multiple sources of information like intensity and edges for finding the pupil center
\item A multistage filtering of candidates is proposed which reduces the error in the final estimate using a scale space approach
\item A simple yet effective pupil tracking scheme is also included for enhanced detection rates and speed.

\end{itemize}

The rest of the paper is organized as follows. Section 2 describes some of the latest works related to localization of pupil in NIR images. Section 3 presents the proposed approach. Extensive testing and evaluations of the proposed approach along with comparisons with the state of the art are presented in Section 4. The advantages and limitations of the proposed methods are discussed in Section 5. Conclusions regarding the method developed along with the future directions of research are discussed in Section 6. 

\section{Related works}

There is a significant amount of work related to pupil localization in NIR images. However, most of the works address the issues in controlled conditions. In this section, we review some of the recent works which discuss robust pupil center localization algorithms.

Li \textit{et al.} \cite{li2005starburst} proposed a hybrid algorithm for pupil center localization combining feature-based and model-based approaches. The algorithm detects and removes the corneal reflections in the preprocessing stage. They detected the pupil
edges by tracing edges along a ray that extends from the best guess pupil center. This method was iteratively used to detect pupil boundary points. RANSAC-based ellipse fitting was carried out using the candidate boundary points. The ellipse parameters thus obtained were used as initial parameters for model-based refinement. 

San Agustin \textit{et al.} \cite{san2010evaluation} proposed a method for eye tracking in low-cost webcam known as `ITU gaze tracker'.  In the first stage, the pupil image is thresholded. The pupil boundary points were detected and fitted to an ellipse using RANSAC based approach. However, the performance of this approach degraded in real-world conditions such as motion blur, glints and reflections.

Swirski \textit{et al.} \cite{swirski2012robust} presented a pupil localization algorithm which can work in highly off-axis images. The coarse location of the pupil was obtained using Haar-like features. From the regions obtained, the pupil was thresholded using K-means clustering method.  A gradient aware RANSAC ellipse fitting was used for fitting the pupil boundary.  The image aware nature of the algorithm made sure that the ellipse boundary lies along strong image edges. However, this method also fails during challenging conditions where external reflections affect the gradients.

Valenti and Gevers \cite{valenti2012accurate} proposed a new method for the localization of iris in visible images. The illumination invariant isophote curvature properties of the edge pixels were used in their approach. The curvatures of edge pixels vote to find the mean iso-center (MIC). This method can be used for IR images as well. However, this method fails to achieve satisfactory accuracy for off-axis images. George \textit{et al.} proposed a method \cite{george2016fast} for iris localization in low resolution images. However, performance in IR images is not studied in the work. 

Kassner \textit{et al.} \cite{kassner2014pupil} proposed an open source framework for gaze tracking using head-mounted cameras along with an open source hardware design. In their approach, pupil candidates were detected using the center surround Haar-like features. A Canny edge detection stage was carried out followed by an edge filtering stage based on neighboring pixels. From the histogram, edges corresponding to spectral reflections were removed. After this edge pruning, remaining edges were labeled using connected components and split into sub-contours based on curvature continuity.  Ellipses were fitted to these candidate contours and evaluated for the inclusion of other contours.  Finally, a confidence score was calculated based on the ratio of supporting edge length and the circumference of the ellipse. If the confidence is less than a threshold, it reports that no ellipse has been found. One of the major disadvantages of this method is that it depends explicitly on edge detection.  If the edge detection stage fails to detect pupil boundary due to motion
blur or illumination, subsequent stages could fail. 

Javadi \textit{et al.} \cite{javadi2015set} proposed SET approach, in which a manual threshold was used for thresholding the image. The connected components were treated as pupil candidates, and their convex hull was found. An ellipse fitting stage was followed, and the ellipse which was closest to the circle shape was selected as the final pupil location. 

Fuhl \textit{et al.} \cite{fuhl2015excuse} presented a robust algorithm for PC localization in off-axis images. In the initial stage, the images were normalized, and the peak of the histogram was found. If the peak was found, the pipeline using edge filtering approach was used. The edges detected by Canny algorithm were filtered using morphological operations for removing lines and orthogonal edges. Straight lines were detected and removed using the distance of points to their centroids. The curve with lowest enclosed intensity was selected and fitted with an ellipse. In case peak was not found, the algorithm finds the coarse location of the pupil and refines it based on angular projection functions. The thresholded image was further refined and fitted with an ellipse. Fuhl \textit{et al.} \cite{fuhl2015else} further extended the work in \cite{fuhl2015excuse} improving it by the use of ellipse selection from Canny edges. After detection of edges using the Canny filter, edge segments were evaluated similarly as used in  ExCuSe \cite{fuhl2015excuse}. The segments were evaluated for various constraints including straightness, the inner intensity value and the best one was fitted. In case ellipse detection fails, likely locations of the pupil were found. The image was downscaled and convolved with a surface difference filter and a mean filter. The location of the maximum value in the multiplied result was used as the initial point for position refinement.  The position is refined based on the analysis of surrounding pixels. The center of mass of the pixels with the new threshold is used as the updated pupil position. This location is evaluated with a validity check using the surface difference.

Most of the methods use multiple stages for PC localization. However, the rather simple assumptions of the pupil as the darkest region produces a lot of false detections. Head-mounted trackers are supposed to work in real-world applications, and they should perform robustly in real-world conditions. To this end, ElSe approach proposed by Fuhl \textit{et al.} \cite{fuhl2015else} is robust. However, their method relies heavily on the Canny edge detector. Once the detector fails to detect the edges correctly due to glints or reflections, the second stage cannot recover if the edge detection fails. Further, they do not leverage the temporal information.

Therefore, we propose a novel method which works even with challenging conditions such as glints, extreme angles, partial occlusion, image blur and illumination variations. Further, we introduce a simple yet effective pupil tracking scheme which makes the detection faster. Usage of the temporal information reduces the search space for pupil localization while decreasing the false positives.

\section{Proposed method}
\begin{figure*}[t]
\begin{center}
\includegraphics[width=1\linewidth]{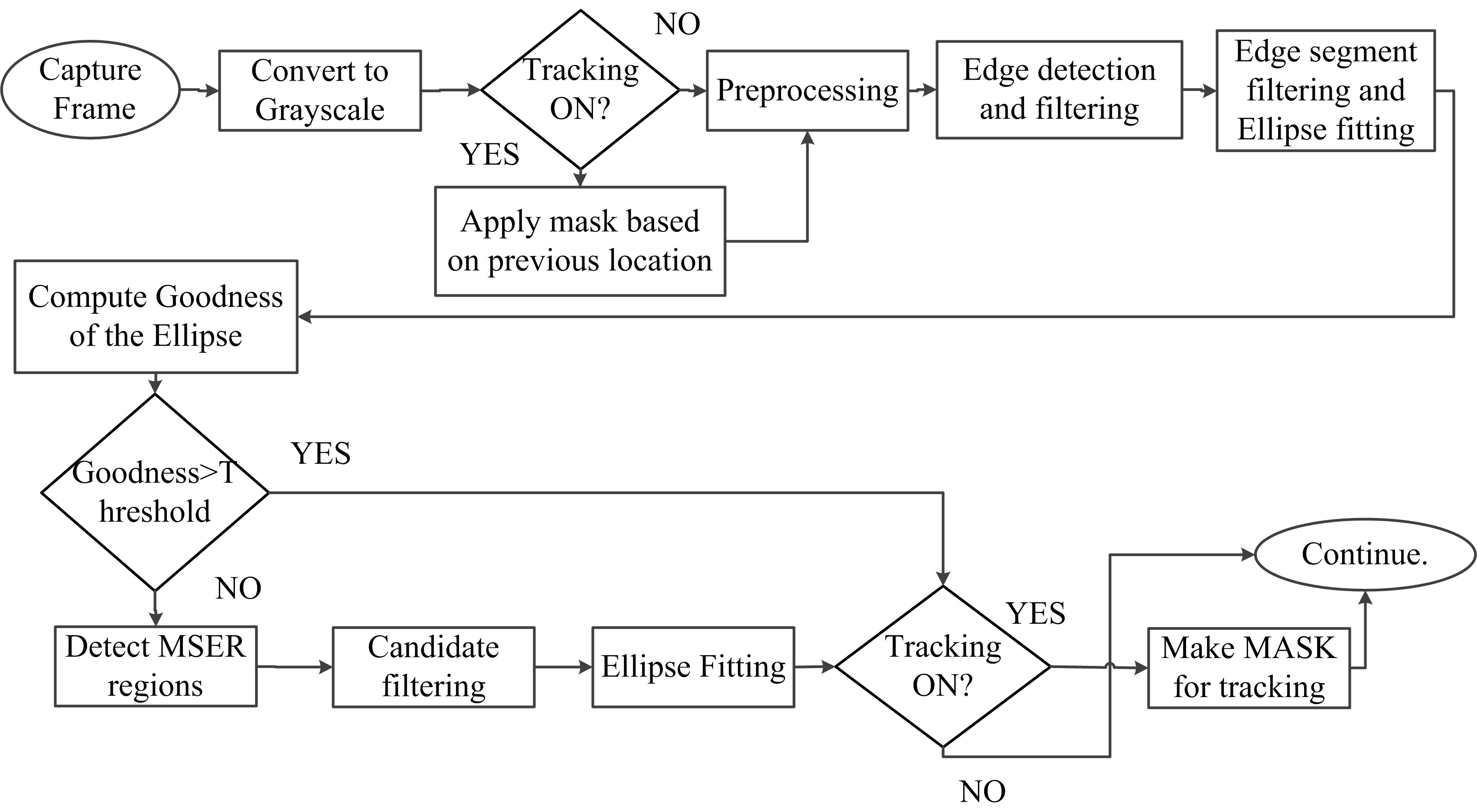}
\end{center}
\caption{Flowchart of the proposed approach}
\label{fig:flowchart}
\end{figure*}

Two basic approaches are commonly used for localizing pupil center in dark pupil images.  The first method uses the intensity distribution of the images. The pupil region is assumed as the darkest region in the image which can be well separated from the background. Some of the approaches use a manual threshold which is adjusted according to the imaging conditions. However, this method fails when other regions appear dark due to the shadows. Further, it may not be possible to find an exact threshold to segment out the pupil due to the varying external lighting and the glints.

Another approach uses edges of the image which can be found using Canny edge detector. The edges detected are filtered morphologically and using several other constraints. Candidate edge segments identified, and ellipse fitting is carried out. However, edge detection stage might fail due to external illumination, glints, and motion blur. In such conditions, the Canny edge detector fails in detecting the pupil boundaries resulting in the failure of subsequent stages.

In our approach, we used intensity distribution, edges, image gradients and several other parameters to estimate the pupil center. The stages of the proposed method and the overall framework is shown in Fig. \ref{fig:flowchart}.

\subsection{Preprocessing and edge detection}
The native resolution of the camera used is 640 $\times$ 480. The images captured are downsampled by a factor of two to reduce the computational requirement. They are converted to grayscale and are scaled to the range of 0-255. 

After obtaining the normalized image, Canny edge detection algorithm is employed for detecting the edges. However, directly applying Canny algorithm over the eye image results in a lot of spurious edges. In our case, the task is to identify the pupil boundary. Since the region inside pupil is somewhat homogeneous, detection of false edges can be reduced by convolving the image with a Gaussian kernel (a $5 \times 5$ kernel was used). This is followed by a median filter stage, which again reduces the number of edges obtained. The Canny algorithm is applied on this preprocessed image, which results in edge segments which are more continuous. This further reduces the computation in the subsequent stages.

\begin{figure*}[h]
\begin{center}
\includegraphics[width=1\linewidth]{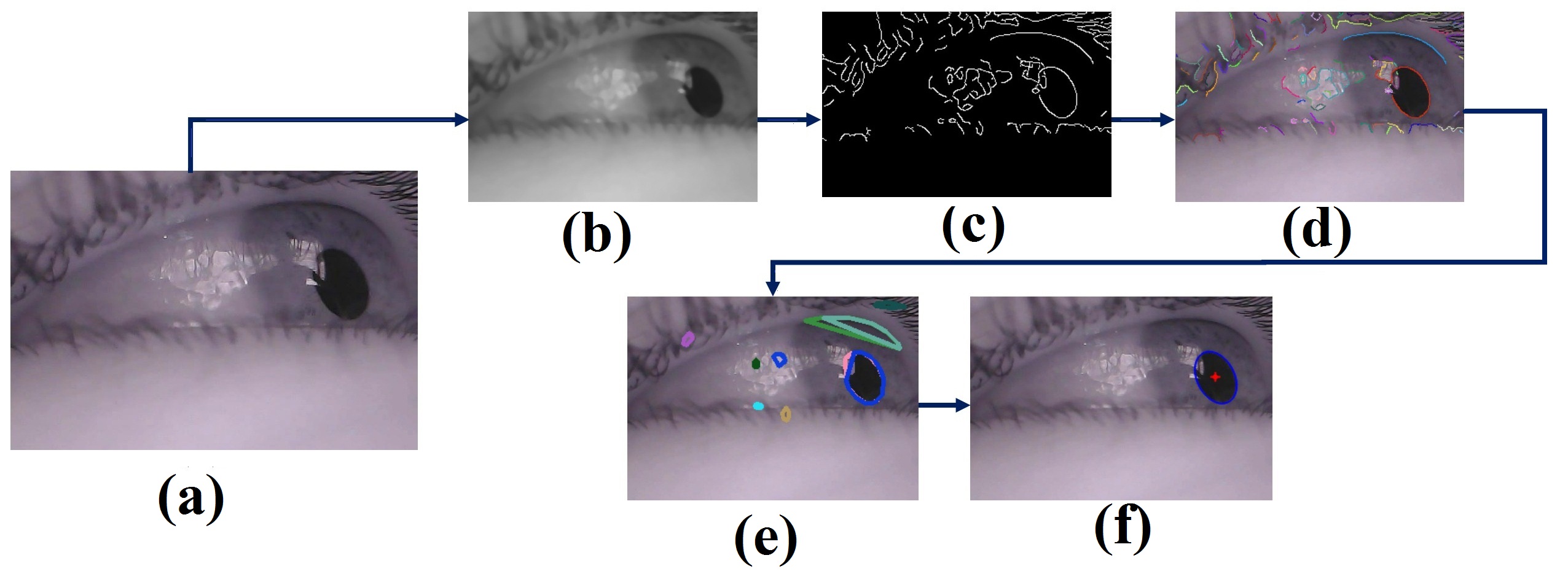}
\end{center}
\caption{Edge based ellipse fitting, a) The original color image captured, b) Downsampled and filtered grayscale image, c) Canny edges, d) Edge segments, e) Candidate edges, f) Fitted ellipse after contour merging }
\label{fig:process_edge}
\end{figure*}

\begin{figure*}[h]
\begin{center}
\includegraphics[width=1\linewidth]{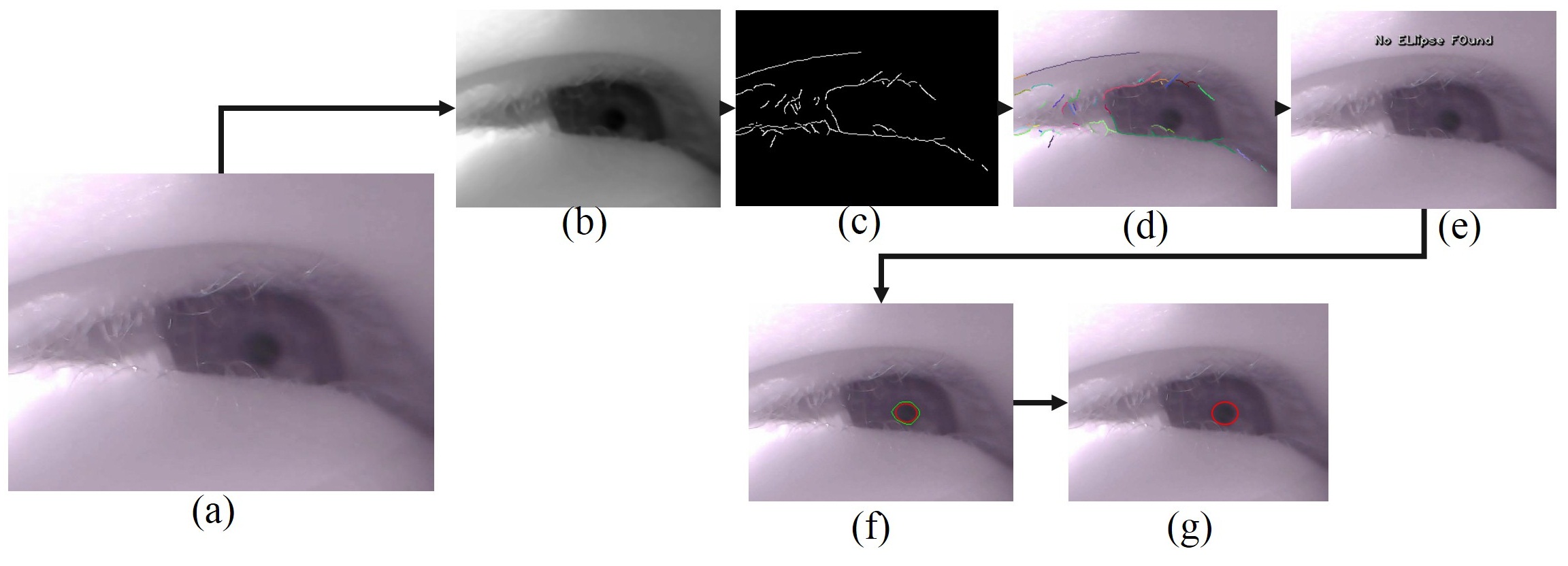}
\end{center}
\vspace{-1.5em}
\caption{Intensity based ellipse fitting, a) The original color image captured, b) Downsampled and filtered grayscale image, c) Detected edges, d) Candidate edge segments, e) Failure of edge based ellipse fitting stage, f) Pruned MSER regions found from the scale space implementation, g) Ellipse fitting corresponding to best pupil candidate }
\label{fig:process_mser}
\end{figure*}

\subsection{Edge selection and candidate filtering}

Once the edges are obtained,  border following algorithm \cite{suzuki1985topological} is used to separate the edge segments. Edge segments with length more than ten are selected for further analysis. Polygonal approximations of the edge segments are found using Douglas-Peucker algorithm \cite{douglas1973algorithms}. Now for each segment, the curvature is computed \cite{kassner2014pupil}, and the segments are split into subsegments if curvature inflections are found. The candidate edge segments are evaluated for the suitability of being a pupil boundary. For this, we introduce new criteria based on ellipse fitting; each edge segment is fitted with an ellipse using a least square approach \cite{Fitzgibbon95abuyer's}.  Candidate edge segments are pruned based on the area and the aspect ratio of the fitted ellipses. Edge segments which are too small or too large are rejected at this stage. The median intensity of the inner region of candidate edge segments are found, and candidates with inner intensity less than an empirically determined threshold are selected for further analysis. We use a new method for candidate edge filtering and merging. The edge segments are sorted based on the median of the inner intensities. In the next stage, the edge candidates belonging to the pupil boundary are merged. A combinatorial search is carried out to determine the whether two candidates belong to the same ellipse. Two parameters are considered in this search, i.e., 1) the similarity of median grayscale value enclosed by the segment, and 2) the Euclidean distance between the centers of the fitted ellipses. Edge segments are merged if these two criteria are satisfied. The combined boundary is fitted with an ellipse, and a goodness parameter is computed. The median difference of grayscale values from the inner and outer, along with the edge support is also computed. We use the goodness of fit parameter proposed in our earlier work \cite{george2016fast}. The center of the ellipse is returned as the pupil center if the goodness parameter is greater than an empirically selected threshold.

 Edge-based ellipse fitting can fail when the edges in the image are weak. Motion blur, low contrast, external lighting, and noise can also fail the edge detection stage. If the edge based fitting fails, we use grayscale intensity distribution to identify the pupil center candidates.

\begin{figure*}[h]
\begin{center}
\includegraphics[width=0.9\linewidth]{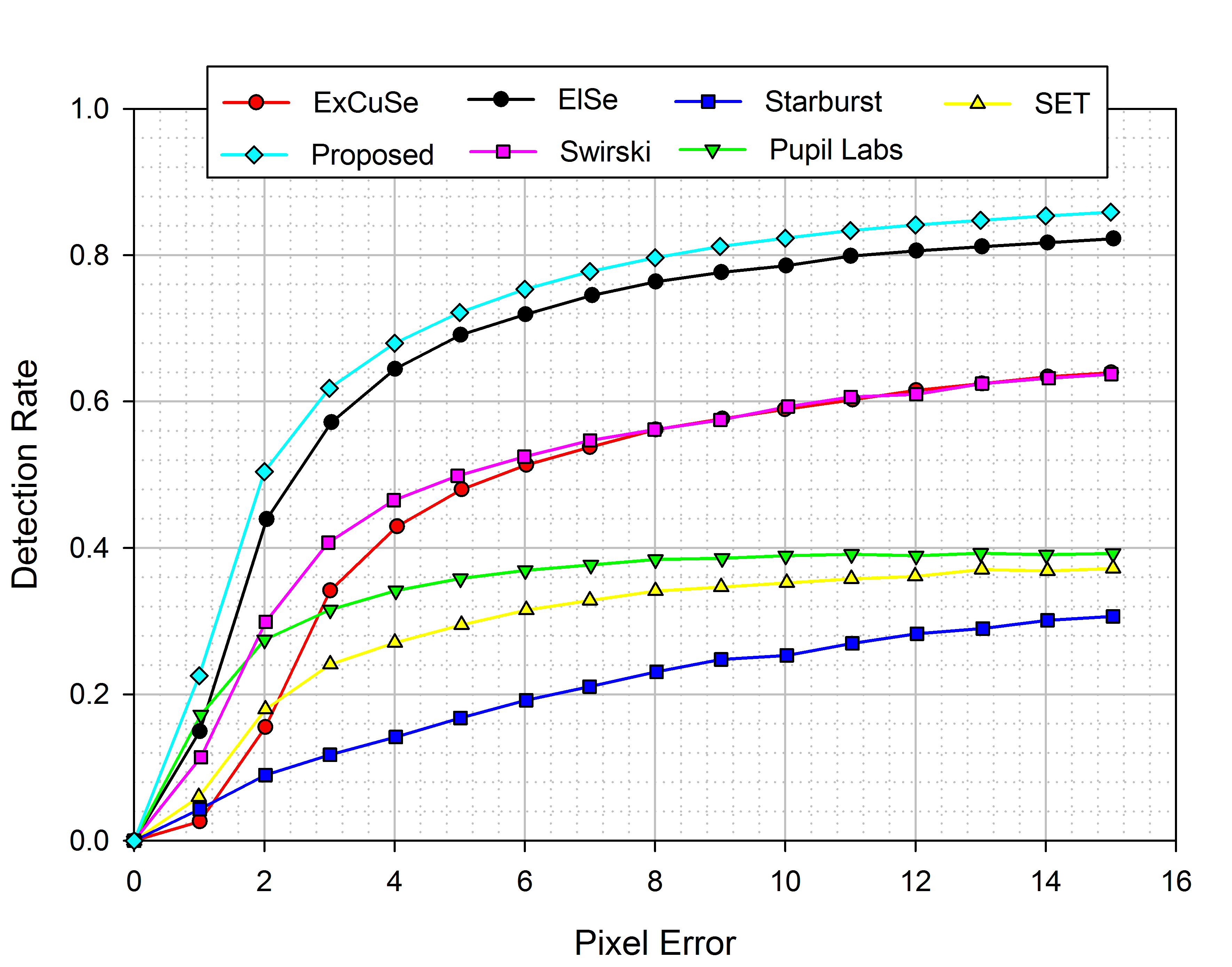}
\end{center}
\vspace{-1.5em}
\caption{Detection rates of the algorithms in LPW dataset ( ElSe, ExCuSe, Pupil Labs, SET, Starburst, Swirski and Proposed).}
\label{fig:lpwfull}
\end{figure*}

\subsection{Candidate detection with MSER}

If the edge-based fitting fails, the algorithm tries to detect the pupil location using the grayscale intensity as shown in Fig. \ref{fig:process_mser}. We apply a candidate filtering approach for obtaining the pupil center. In the first stage, different candidate regions are identified using a variant of maximally stable extremal regions (MSER) proposed by Matas \textit{et al.} \cite{matas2004robust}. In the second stage, the candidate regions are evaluated for ellipse fitting criterion, and the best candidate is selected as the pupil center.

Component tree of an image is the set containing all the connected components of different thresholds, ordered by inclusion. The threshold starts from zero to the highest value of 255. Connected components change as the threshold for segmentation is varied from 0-255. The maximally stable extremal regions can be found from the component tree of a grayscale image. We start with the lowest value in the image grayscale values. Connected components with different thresholds are found out. The region corresponding to a particular threshold is said to be stable if the area of the thresholded regions remains almost stable over a large range of thresholds.  The local maxima of these regions are identified as the maximally stable extremal regions. Even though the direct implementation is computationally complex, a linear time in the number of pixel solution is available in \cite{nister2008linear}. More details about the fast implementation of MSER algorithm can be found in \cite{nister2008linear}. In our approach, we use three constraints in detecting the MSERs. The minimum and maximum areas of the pupil are assumed to be known, which are used as constraints. The maximum inner intensity of the pupil region is also known. The component tree needs to be computed only up to this level for finding the candidate regions. MSER is known to be sensitive to image blur. Scale-space pyramid based implementation \cite{forssen2007shape} is used to alleviate the issue of image blur. Instead of detecting MSER in a single resolution, a scale-space pyramid is constructed, and MSERs is identified separately in each octave. After detection, duplicate detections are moved considering the size, location, and scale in the scale-space pyramid. This multi-scale approach improves the robustness even in the presence of blur.

Once the MSER regions corresponding to the constraints are obtained, a candidate filtering approach is performed to identify the best pupil candidate. The region boundary contours are fitted with ellipses, and the ratio of the major axis and the minor axis is found. Candidate regions with the ratio less than a 
predefined threshold are identified for further analysis. The goodness parameter is computed, and the candidate with the highest goodness parameter is used as the pupil ellipse.


\subsection{Tracking framework}

The time taken for processing can be reduced significantly by using temporal information. Tracking algorithms like Kalman filter or Particle filter usually assume a motion model for the dynamics of the object to be tracked. However, the dynamics of eye movements involve several subclasses like fixations, saccades, vergence, smooth pursuits, etc. Having different dynamics makes it difficult to implement tracking in practical scenarios. However, based on eye physiology and sampling frequency of the image acquisition system, a rough estimate of the maximum possible change in the position of the pupil center between successive frames can be computed. This information can be used to constrain the search space without the loss of accuracy. In our approach, we have used the previous location of PC to obtain the search region for the current frame. A rectangular region is selected around the last position of the pupil. Search space for PC localization is limited to this area only. However, selection of this mask depends on the confidence of PC localization, the mask from the previous frame is used only when the goodness parameter for ellipse fitting is more than an empirically found threshold. If the ellipse fitting stage in the previous frame does not result in a high Goodness factor, the entire image is searched for localizing PC. This simple approach achieves better frame rates while removing many false detections. The main advantage of this method is that it gets rid of redundant computations and limits the processing to more promising areas using the temporal information.

\subsection{Comparison with state of the art method}

The closest method related to the proposed approach is ElSe \cite{fuhl2015else} proposed by Fuhl \textit{et al.}, as they also use two different pipelines based on the imaging conditions. In the first stage, they have used Canny edge detector followed by complex algorithmic and morphological filtering, whereas in the proposed approach most spurious edges are rejected by downsampling followed by Gaussian filtering. We have introduced new criteria for edge filtering based on the geometric distance and the inner intensity difference of the fitted ellipses. In the event of failure of the edge based stage, the second stage is performed which uses scale space variant of MSER algorithm followed by the proposed candidate filtering to identify the best pupil candidate. Further, the proposed method introduces a tracking approach
which reduces the search space based on the confidence levels obtained from the
ellipse fitting stage.

\section{Experiments}

\begin{table*}[htb]
\centering
\caption{Comparison of detection rates for an error of 5 pixels}
\label{tab:fivepixel}
\begin{tabular}{@{}llllllll@{}}
\toprule
Data set\textsuperscript{*} & \multicolumn{1}{c}{SET (\%)} & \multicolumn{1}{c}{Starburst (\%)} & \multicolumn{1}{c}{Swirski (\%)} & \multicolumn{1}{c}{ExCuSe (\%)} & \multicolumn{1}{c}{ElSe (\%)} & \multicolumn{1}{c}{Pupil Labs (\%)} & \multicolumn{1}{c}{Proposed (\%)} \\ \midrule
1        & 56.86                        & 39.79                              & 84.48                            & 63.53                           & 87.95                         & 65.58                               & \textbf{92.05}                    \\
2        & 48.68                        & 19.70                              & 41.58                            & 29.90                           & 69.87                         & 24.72                               & \textbf{82.83}                    \\
3        & 27.55                        & 6.75                               & 31.43                            & 34.83                           & \textbf{57.50}                & 21.82                               & 51.58                             \\
4        & 7.70                         & 9.27                               & 16.87                            & 25.38                           & 37.42                         & 14.53                               & \textbf{50.92}                    \\
5        & 6.75                         & 0.00                               & 8.38                             & 19.08                           & \textbf{22.95}                & 13.15                               & 19.75                             \\
6        & 11.10                        & 13.30                              & 63.48                            & 53.44                           & 84.10                         & 38.72                               & \textbf{87.91}                    \\
7        & 43.55                        & 7.65                               & 66.17                            & 66.48                           & 73.60                         & 68.43                               & \textbf{85.93}                    \\
8        & 42.17                        & 34.32                              & 77.68                            & 75.32                           & 81.00                         & 64.22                               & \textbf{87.25}                    \\
9        & 35.65                        & 30.90                              & 56.40                            & 60.42                           & \textbf{61.97}                & 45.20                               & 61.53                             \\
10       & 10.42                        & 3.65                               & 71.23                            & 59.00                           & 72.65                         & 41.62                               & \textbf{82.93}                    \\
11       & 31.07                        & 18.10                              & 31.58                            & 49.52                           & 71.48                         & 9.45                                & \textbf{73.68}                   \\
12       & 54.92                        & 24.10                              & 71.82                            & 72.58                           & \textbf{89.73}                & 49.58                               & 89.32                             \\
13       & 14.75                        & 16.52                              & 27.03                            & 45.04                           & \textbf{51.51}                & 16.68                               & 45.43                             \\
14       & 30.25                        & 23.50                              & 76.07                            & 58.60                           & 70.50                         & 57.52                               & \textbf{79.40}                    \\
15       & 27.17                        & 8.15                               & 37.80                            & 44.83                           & 53.95                         & 43.78                               & \textbf{58.48}                    \\
16       & 23.24                        & 17.24                              & 74.11                            & 72.73                           & 82.13                         & 82.01                               & \textbf{85.67}                    \\
17       & 20.90                        & 2.42                               & 68.88                            & 42.10                           & 72.97                         & 47.52                               & \textbf{73.15}                    \\
18       & 50.67                        & 33.48                              & 61.18                            & 66.25                           & 78.57                         & 48.73                               & \textbf{82.48}                    \\
19       & 11.97                        & 3.45                               & 24.87                            & 21.88                           & 54.05                         & 2.60                                & \textbf{76.72}                    \\
20       & 19.83                        & 16.58                              & 41.40                            & 11.72                           & \textbf{83.35}                & 0.98                                & 76.38                             \\
21       & 30.43                        & 25.63                              & 55.90                            & 47.45                           & 88.92                         & 28.18                               & \textbf{92.05}                    \\
22       & 41.60                        & 11.85                              & 6.48                             & 31.23                           & 70.00                         & 0.83                                & \textbf{75.65}                    \\ \hline
Overall  & 29.42                        & 16.65                              & 49.76                            & 47.79                           & 68.92                         & 35.72                               & \textbf{73.23}                   \\
\bottomrule
\multicolumn{8}{l}{\textsuperscript{*}\footnotesize{Data taken from \cite{fuhl2016pupil} for comparison.}}\\

\multicolumn{8}{l}{\footnotesize{Best results obtained in each dataset is shown in bold scripts.}}

\end{tabular}
\end{table*}

\subsection{Labeled Pupils in the Wild database}
We have used labeled pupils in the wild (LPW) database \cite{tonsen2016labelled} for the algorithm evaluation since the number of images is large and it contains images recorded in real-world conditions.  LPW dataset contains 66 high-quality videos of eye regions from 22 subjects, including samples from people of different ethnicities, indoor and outdoor illumination variations in different gaze directions. It also contains images of participants wearing glasses, contact lenses and makeup. Each video in the database contains around 2000 images of resolution $640 \times 480$ recorded at a frame rate of 95 fps. The dataset contains a total of 130,856  images which is much larger than any of the existing datasets. Ground truth pupil locations are also available with the dataset.


\subsection{Evaluation of the algorithm}

We have evaluated the proposed algorithm in 66 videos provided in the dataset. The pixel error in each frame was computed from the ground truth available with the dataset. The comparison with state of the art is made based on the data as in \cite{fuhl2016pupil}.

The result obtained from the proposed algorithm has been compared with six other state of the art methods available in the literature. We have compared the results with Starburst \cite{li2005starburst}, Swirski \cite{swirski2012robust}, SET \cite{javadi2015set}, Pupil Labs \cite{kassner2014pupil}, ExCuSE \cite{fuhl2015excuse}, and ElSE \cite{fuhl2015else}.

The results obtained, and the comparison with state of the art is shown in Fig. \ref{fig:lpwfull}. The proposed approach outperforms all the state of the art methods. Comparative results for an error of five pixels is provided in Table. \ref{tab:fivepixel}. The proposed method obtains best overall accuracy.

\begin{figure}[h]
\begin{center}
\includegraphics[width=1\linewidth]{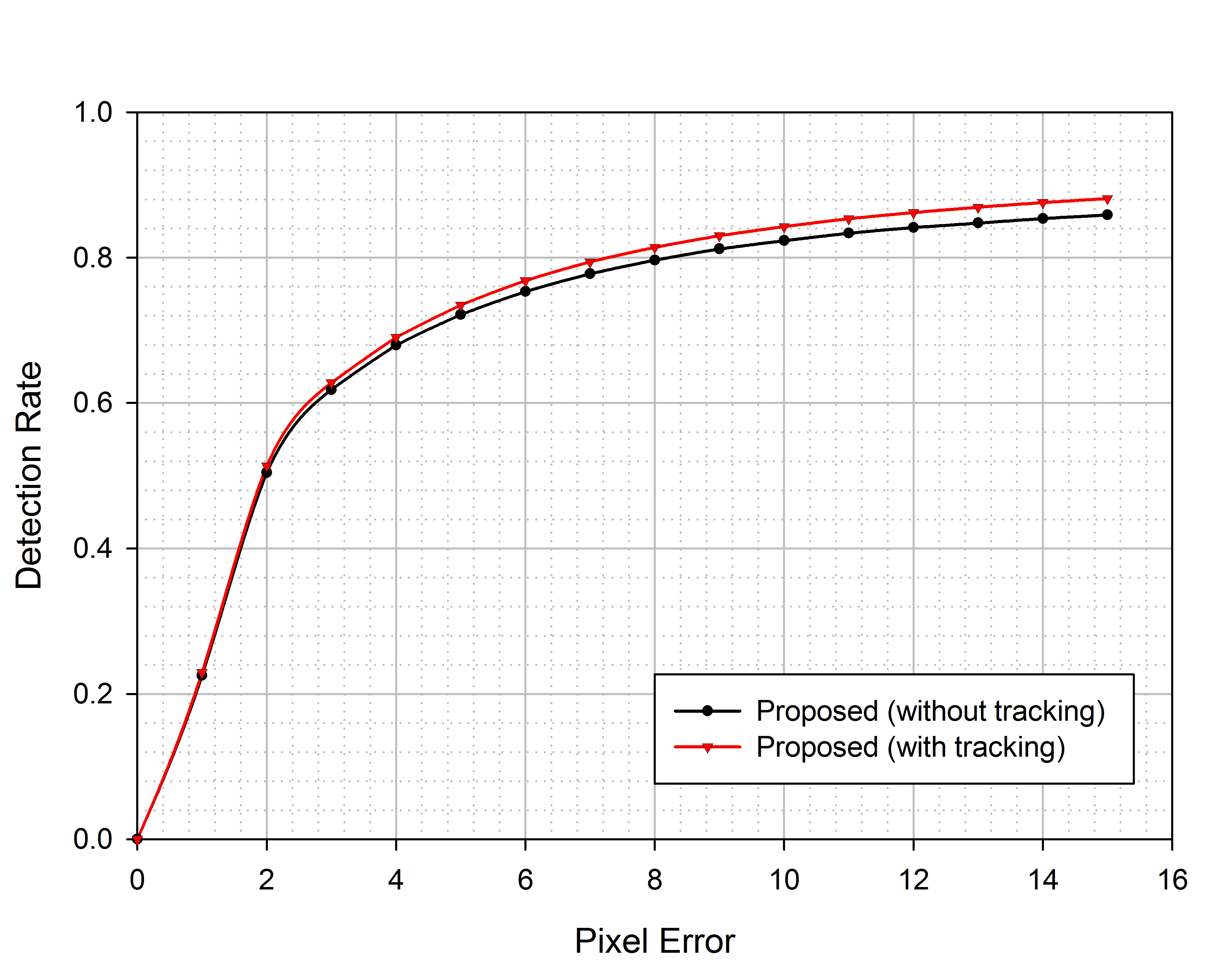}
\end{center}
\caption{Detection rates with and without tracking.}
\label{fig:tracking}
\end{figure}

The results obtained with and without tracking are shown in Fig. \ref{fig:tracking}. The addition of tracking decreased the processing load without any decrease in accuracy. The results obtained with tracking are slightly better than individual frame based detections. This can be attributed to the reduction in false detections due to the masking used in the tracking approach. 
\begin{figure}[h]
\begin{center}
\includegraphics[width=1\linewidth]{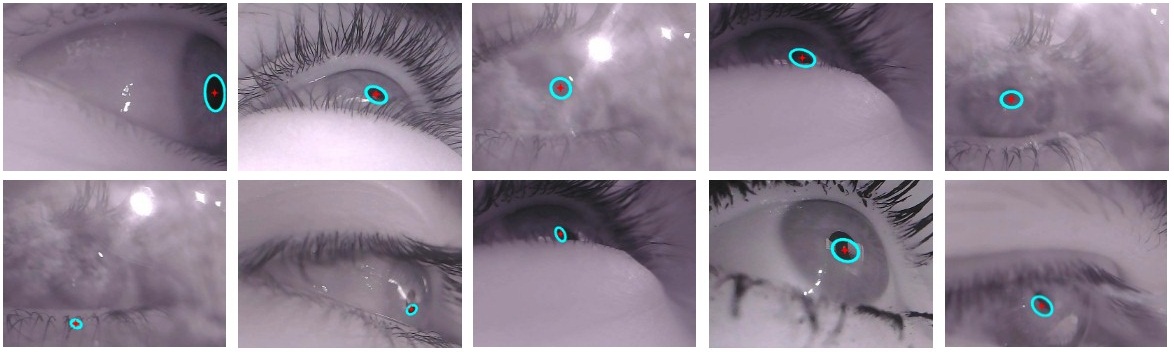}
\end{center}
\caption{Sample results from the detections, the first row shows the successful detection and second row shows detection failures.}
\vspace{-.5em}
\label{fig:success_failure}
\end{figure}
\begin{figure}[h]
\begin{center}
\includegraphics[width=1\linewidth]{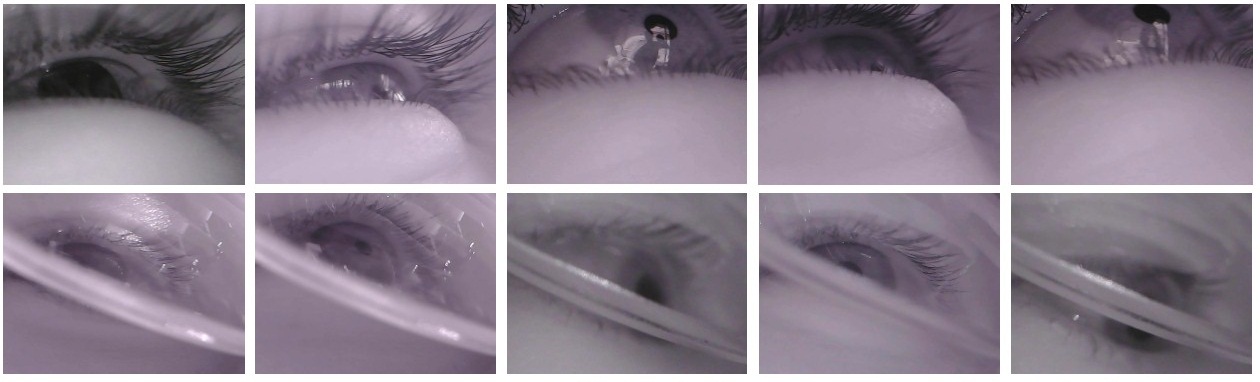}
\end{center}
\caption{Some examples of the challenging images from datasets 4 and 5}
\vspace{-1.5em}
\label{fig:challenging}
\end{figure}

Some of the successful detections and failures are shown in Fig. \ref{fig:success_failure}. Fig. \ref{fig:challenging} shows some of the challenging images from the LPW dataset.

%
%

\section{Discussions}
The overall performance of the algorithm in the entire dataset is shown in Fig. \ref{fig:lpwfull}. The proposed algorithm outperformed all the state of the algorithms.  

The algorithm was designed to be robust against real-world conditions. Detection using one particular feature may not work in all practical usage scenarios.
The proposed algorithm switches to either edge based method or intensity based
method depending upon the image conditions. One significant advantage here
is that, even if the first edge based stage of the algorithm fails due to some reflections, the second stage can identify the pupil (though more computation is
required). Further, the tracking approach reduces false detection rate at the same time reduces computational load as the search space is considerably less. Most of the algorithms are designed to maximize the per frame based detection rates. Here, we have added the tracking framework which directly extends the algorithm for video. Results with and without tracking are shown in Fig. \ref{fig:tracking}. The tracking scheme achieved slightly better results with better runtime performance.

\subsection{Execution time }

The algorithms were implemented on a desktop computer with 64 bit Ubuntu 13.10 Operating system having  3.33 GHz core i5 processor, and  8GB RAM. Implementation with unoptimized python code obtained an average processing time of 14.28 ms/frame without tracking and 9.90 ms/frame with tracking.  Execution time comparison for four best systems is shown below in Table. \ref{tab:exectime}. The value reported is the average time for detection in the database. However, the execution time will be less for images where the edge-based approach succeeds. In the worst case scenario, where the edge detection stage fails, the processing time could be high for those particular images (since both edge and MSER pipelines need to run in this case). There is an opportunity to skip the edge based pipeline if the edge stage is failing for continuous frames. The execution time reported here is more close to real-world conditions as it is the average from a large, in the wild database. It is to be noted that the current implementation is in python. There is a scope for improving the processing time by code optimization, and implementing computationally complex part in C++.

\begin{table}[]
\centering
\caption{Execution time comparison}
\label{tab:exectime}
\begin{tabular}{@{}ll@{}}
\toprule
         & Execution time per image (in ms) \\ \midrule
Proposed & 9.9                              \\
Else     & 7                                \\
Swirski  & 8                                \\
Excuse   & 6                                \\ \bottomrule
\end{tabular}
\end{table}

\subsection{Limitations}

The algorithm detects the pupil centers accurately when the edges are correctly detected. Intensity-based candidate filtering approach detects the pupil when the edge-based approach fails. However, the algorithm fails when the pupil is occluded as shown in Fig. \ref{fig:challenging} (dataset 5). The edges are not properly detected because of the blur. Intensity-based approach could also fail since the candidates are occluded. This reduces the goodness of regions obtained. Another failure case can occur when the image contrast is poor, and the surrounding regions have low contrast and reflections. False positives with very high confidence score in the detection stage can result in incorrect detection in the subsequent frame because of the tracking scheme.  A machine learning based approach can be used to compensate for the detection-failures in such challenging cases.


\section{Conclusions}

In this work, we have developed a framework for pupil centre detection in IR images. The algorithm works in images captured using head-mounted eye tracker using dark pupil method. The main objective of the work was to develop an accurate algorithm which would work in real-world conditions. The algorithm takes care of both intensity and edge information to estimate the pupil centre accurately. A candidate filtering approach is chosen which maximizes a goodness function returning the best possible pupil candidate. A simple tracking approach has also been used, which reduces the computations required without the reduction in accuracy. The proposed approach has been evaluated on Labelled Pupil in the Wild dataset and found to outperform all the state of the art methods.  The Python-based implementation achieves frame rates close to 100, which can be improved with an optimized implementation in C/C++. The high frame rates achieved can be useful in adding additional post-processing and more computationally complex tracking algorithms for improving the accuracy even further. Online identification of the eye movement types can be helpful in using appropriate model for tracking eye movements during different movement types. The addition of machine learning methods to improve the challenging conditions is another path to be explored.

\section*{Acknowledgements}
The authors would like to thank researchers from Max Planck Institute for Informatics for providing the database.

\bibliographystyle{IEEEtran}

\bibliography{refs_ireye}

\end{document}